\documentclass[10pt,twocolumn,letterpaper]{article}

\usepackage[pagenumbers]{cvpr} %

\usepackage{booktabs}
\usepackage{multirow}
\usepackage{rotating}
\usepackage{adjustbox}
\usepackage{bm}
\usepackage{soul}
\def\eg{\emph{e.g}\onedot}

\def\ie{\emph{i.e}\onedot}

\definecolor{cvprblue}{rgb}{0.21,0.49,0.74}
\usepackage[pagebackref,breaklinks,colorlinks,allcolors=cvprblue]{hyperref}

\def\paperID{*****} %
\def\confName{CVPR}
\def\confYear{2025}

\title{All You Need to Know About Training Image Retrieval Models}

\author{
Gabriele Berton\\
Polytechnic of Turin\\
\and
Kevin Musgrave\\
Setta.dev\\
\and
Carlo Masone\\
Polytechnic of Turin\\
}

\begin{document}
\maketitle

\begin{abstract}

Image retrieval is the task of finding images in a database that are most similar to a given query image. The performance of an image retrieval pipeline depends on many training-time factors, including the embedding model architecture, loss function, data sampler, mining function, learning rate(s), and batch size. In this work, we run tens of thousands of training runs to understand the effect each of these factors has on retrieval accuracy.
We also discover best practices that hold across multiple datasets.
The code is available at
{\small{\url{https://github.com/gmberton/image-retrieval}}}
\end{abstract}

\section{Introduction}
\label{sec:introduction}

Image retrieval systems form the backbone of many applications we use daily, from visual search engines to content-based product recommendations. This report serves as a resource for understanding the factors that influence image retrieval model training. Using a consistent methodology, we address practical questions that developers face, such as:
\begin{itemize}
    \item Which layers of the base model should be fine-tuned?
    \item How should the learning rates be set?
    \item How should the training dataset be sampled?
    \item When creating a dataset, should the main focus be annotation quality, or dataset size?
    \item What feature layer and feature dimensionality result in the best accuracy?
\end{itemize}

\subsection{Findings}

\paragraph{Model, Optimizer, \& Learning Rate}
Our experimental results show that an ideal image retrieval pipeline relies on the CLS features of DINO-v2 \cite{Oquab2023DINOv2LR}, tuning all layers with Adam \cite{Kingma2014AdamAM} and a learning rate of 1e-6.

\paragraph{High-Resource Setting}
When substantial GPU resources are available, contrastive losses (such as Threshold-Consistent Margin \cite{Zhang2023ThresholdConsistentML} and Multi-Similarity \cite{Wang2019MultiSimilarityLW}) combined with an online miner yield superior results, as they benefit from larger batch sizes. Additionally, our findings indicate that sampling a limited number of images (2-4) per class within each training batch produces the best results.

\paragraph{Resource-Constrained Setting}
When GPU memory is limited (\eg batch size is lower than 256), classification losses, like CosFace \cite{Wang2018CosFaceLM} and ArcFace \cite{Deng2018ArcFaceAA}, fare much better. These benefit from using a single image per class within each batch. Additionally, for the loss's classifier layer, the learning rate should be much higher than the learning rate used to fine-tune the model (\eg 1 or higher, versus an ideal learning rate of 1e-6 to fine-tune the model).

\paragraph{Dataset Labeling Strategy}
Finally, for anyone labeling a dataset, we find that all metric learning losses are robust to a few wrong annotations, while all benefit from using a bigger training dataset (\eg more classes). Hence, one should focus on labeling more data rather than labeling data very carefully.\\

\noindent These findings represent practical guidelines derived from systematic experimentation across multiple datasets. Practitioners can use these insights to make informed decisions when designing and optimizing their image retrieval systems. In the remainder of this paper, we discuss relevant prior work (Section \ref{sec:related_work}) and present our experimental methodology and results (Section \ref{sec:experiments}), with subsections examining various factors affecting retrieval performance.

\section{Related Work}
\label{sec:related_work}

Image retrieval is a widely studied task in computer vision, typically addressed through deep metric learning methods, where the objective is to learn embeddings that capture visual similarity effectively. Central to the success of these methods are loss functions, which guide the training of embeddings. These loss functions can be broadly categorized into two main approaches: contrastive losses, which directly optimize the distances between groups of embeddings, and classification losses, which learn discriminative embeddings via class prediction. Since classification losses require extra learnable parameters, and contrastive losses do not, these two categories can also be thought of as ``stateful" (classification) and ``stateless" (contrastive), as discussed in \cref{tab:contrastive_loss_comparison}.

While multiple benchmarking studies exist in this domain \cite{Fehrvri2019UnbiasedEO, Roth2020RevisitingTS, Musgrave_2020_MLRC, Milbich2021CharacterizingGU}, they usually focus on comparing different losses and miners, without exploring how each component interacts with the others.
On the other hand, this paper provides a careful study on each one of the most important pieces that make up a retrieval pipeline, presenting insights derived from thousands of systematic training runs.

\begin{table}[h]
\begin{center}
\begin{adjustbox}{width=0.99\linewidth}
\begin{tabular}{c|c}
\toprule
\textbf{Contrastive or Stateless} & \textbf{Classification or Stateful} \\
\midrule
Examples: Triplet Loss, NT-Xent & Examples: CosFace, ArcFace\\
\midrule
No extra classifier & Uses an extra classifier (only at training time) \\
No trainable parameters & Uses extra trainable parameters for the classifier \\
- & Important to tune the classifier's LR separately \\
Miner needed (improves results) & No miner needed \\
GPU parallelization is hard & GPU parallelization is easy \\
Larger batch sizes preferred & Works well even with smaller batch sizes \\
- & Requires global consistency for classes* \\
\bottomrule
\end{tabular}
\end{adjustbox}
\caption{
\textbf{Comparison of Stateless and Stateful Loss Functions.}
*Classification losses require each dataset sample to maintain exactly one consistent label throughout training. Contrastive losses have no such requirement, allowing a sample's label to vary between batches. This distinction has important implications for certain tasks. For example, classification losses cannot be used effectively in self-supervised learning, where image views/augmentations receive temporary labels that exist only within the current batch.
}
\label{tab:contrastive_loss_comparison}
\end{center}
\end{table}

\section{Experiments}
\label{sec:experiments}

In this section, we discuss the datasets, methodology, and experiment results.

\subsection{Datasets}
\label{sec:datasets}

\begin{figure*}
    \begin{center}
    \includegraphics[width=0.99\linewidth]{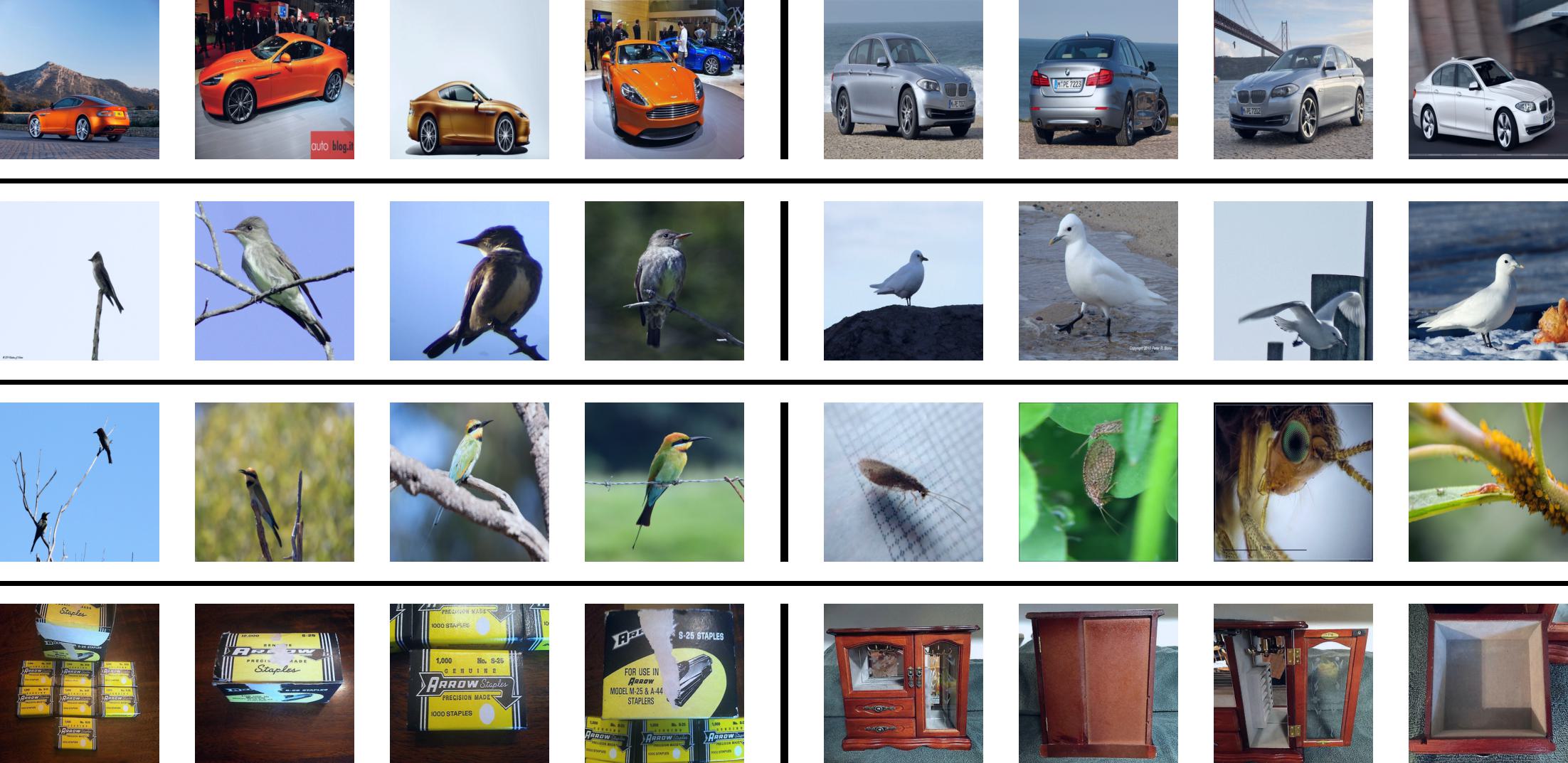}
    \end{center}
    \caption{Examples of images from the four datasets. Each row is a dataset, and each group of four images is a class (two classes per dataset). From top to bottom, the rows are Cars196, CUB, iNaturalist2018, StanfordOnlineProducts.}
    \label{fig:datasets_examples}
\end{figure*}

\begin{figure}
    \begin{center}
    \includegraphics[width=0.99\columnwidth]{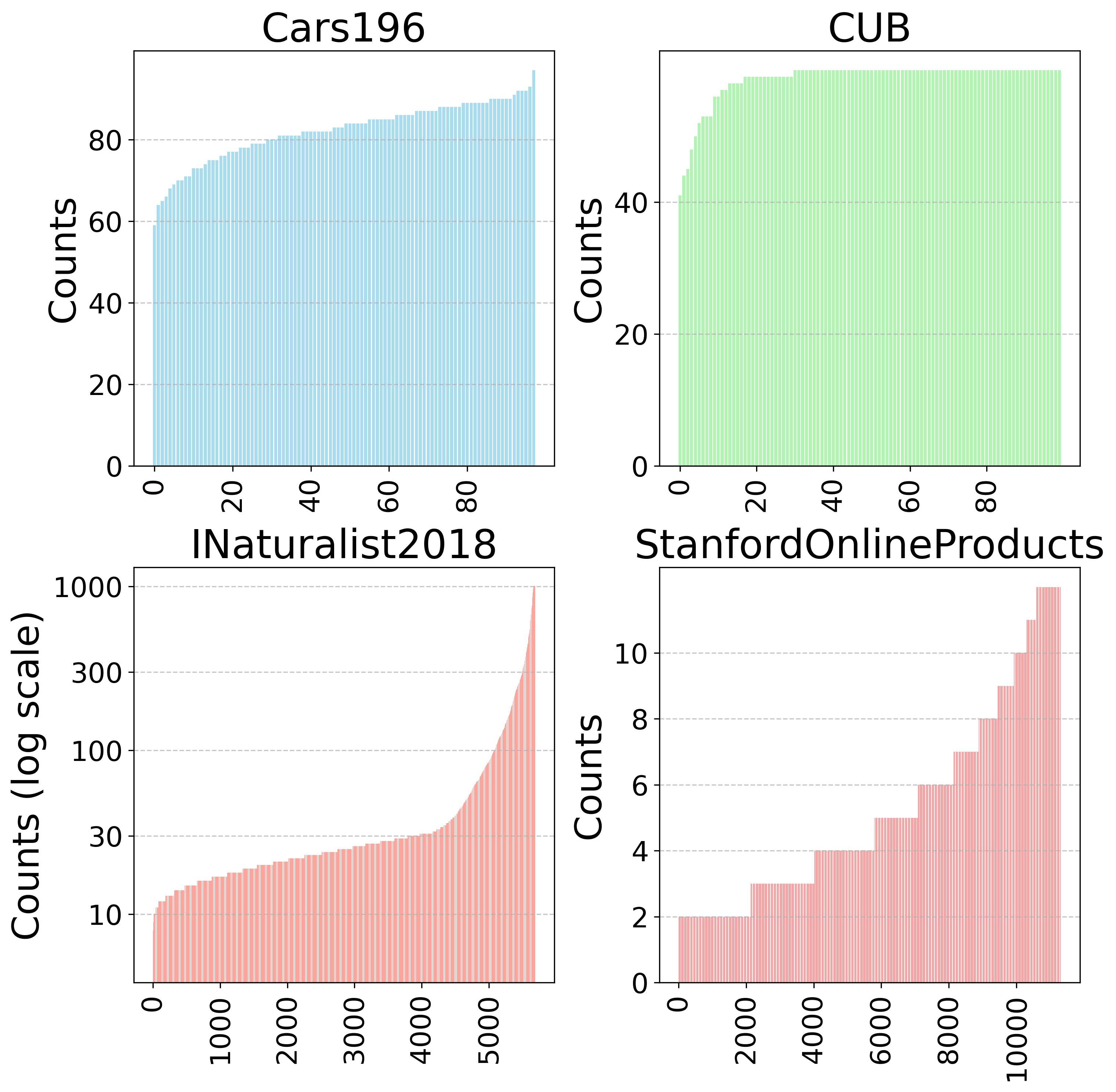}
    \end{center}
    \caption{The number of samples per class within the training set of each dataset. The y-axis for iNaturalist2018 is in log scale.}
    \label{fig:datasets_counters}
\end{figure}

In this benchmark we use four datasets: Cars196 \cite{Dehghan_2017_Cars196}, CUB-200-2011 (CUB) \cite{Wah_2011_CUB}, iNaturalist 2018 (iNaturalist) \cite{Horn_2017_INaturalist}, and Stanford Online Products (SOP) \cite{Song_2016_SOP}. As shown in \cref{fig:datasets_examples} and \cref{fig:datasets_counters}, Cars and CUB are simpler, class-balanced datasets. In contrast, iNaturalist and SOP are more challenging and unbalanced. Table \ref{tab:datasets_splits} contains a summary of the train/test splits of each dataset.

\begin{table*}
\begin{center}
\begin{adjustbox}{width=0.7\linewidth}
\begin{tabular}{c|ccccc}
\toprule
\textbf{Dataset} & \textbf{Train Images} & \textbf{Test Images} & \textbf{Train Classes} & \textbf{Test Classes} & \textbf{Image Types} \\
\midrule
Cars196 & 8054 & 8131 & 98 & 98 & Birds \\
CUB & 5864 & 5924 & 100 & 100 & Cars \\
iNaturalist2018 & 325846 & 136093 & 5690 & 2452 & Species \\
StanfordOnlineProducts & 59551 & 60502 & 11318 & 11316 & Products \\
\bottomrule
\end{tabular}
\end{adjustbox}
\caption{Summary of train/test splits for the datasets used in this paper.}
\label{tab:datasets_splits}
\end{center}
\end{table*}

\subsection{Methodology}

For all our experiments we used the following settings, to ensure a level playing field and reproducibility:

\begin{itemize}
\item{Resize images to 224$\times$224.}
\item{Apply RandAugment \cite{Cubuk_2020_RandAugment} to every batch of images.}
\item{Use the dataset splits specified in the PyTorch Metric Learning (PML) library \cite{Musgrave_2020_PyTorchML}. Each dataset is split into a train-val and test set, and unless otherwise specified, we use 80\% of the train-val split for training and 20\% for validation.}
\item{Train the model for a maximum of 100 epochs, with early stopping if the precision on the validation set has not improved for 3 epochs. In practice, almost all experiments were early stopped within less than 20 epochs.}
\item{Run each experiment three times with different seeds, and report the average result.}
\end{itemize}

\subsection{Findings of our preliminary experiments}
\label{sec:preliminary_experiments}
First, we ran a large number of preliminary experiments to establish good baseline settings for our subsequent in-depth experiments. Here are those baseline settings:

\begin{itemize}
    \item Use the \textbf{DINO-v2-base model} \cite{Oquab2023DINOv2LR}, which outperforms various ResNets and DINO-v2-small models.
    \item Use the \textbf{Adam optimizer}  \cite{Kingma2014AdamAM}, which on average outperforms RMSProp \cite{Tieleman_2012_RMSProp}.
    \item \textbf{Fine-tune the entire model}, instead of fine-tuning only the last layers. For example, fine-tuning the entire DINO-v2 model works better than fine-tuning only the last four transformer blocks.
    \item Use a \textbf{1e-6 learning rate} for the entire model, except for the classifier layer, which is present only for classification losses.
    \item Use a \textbf{classifier layer learning rate of 1} (only for classification losses).
    \item For contrastive losses: \textbf{always use a miner} and a small number of images (\eg 4) per class within each batch.
    \item For classification losses: use \textbf{1 image per class} within each batch.
\end{itemize}

\noindent In our preliminary experiments, we tried all 34 loss functions implemented in PML, and ran a grid search on learning rates for each one. Then we selected the top 12 performing loss functions, which are:

\begin{itemize}
    \item ArcFace \cite{Deng2018ArcFaceAA}
    \item Circle \cite{Sun2020CircleLA}
    \item Contrastive \cite{Hadsell2006DimensionalityRB}
    \item CosFace \cite{Wang2018CosFaceLM}
    \item MultiSimilarity \cite{Wang2019MultiSimilarityLW}
    \item NTXent \cite{Sohn2016ImprovedDM}
    \item NormalizedSoftmax \cite{Zhai2018ClassificationIA}
    \item ProxyAnchor \cite{Kim2020ProxyAL}
    \item SignalToNoiseRatioContrastive \cite{Yuan2019SignalToNoiseRA}
    \item SoftTriple \cite{Qian2019SoftTripleLD}
    \item ThresholdConsistentMargin \cite{Zhang2023ThresholdConsistentML}
    \item TripletMargin \cite{Weinberger2005DistanceML}
\end{itemize}

 We found that for each loss, the default hyperparameters used by PML are near optimal in almost all cases, and therefore we ran our experiments without changing any loss hyperparameters.

Throughout the next sections, we plot results for each of the four datasets using the MAP@R evaluation metric \cite{Musgrave_2020_MLRC}.
Considering the two broad types of metric learning losses, namely contrastive and classification, the upcoming plots show this distinction by having a solid line for contrastive losses and a dotted line for classification losses.

\subsection{Batch size}

\begin{figure}[h]
    \begin{center}
    \includegraphics[width=0.99\linewidth]{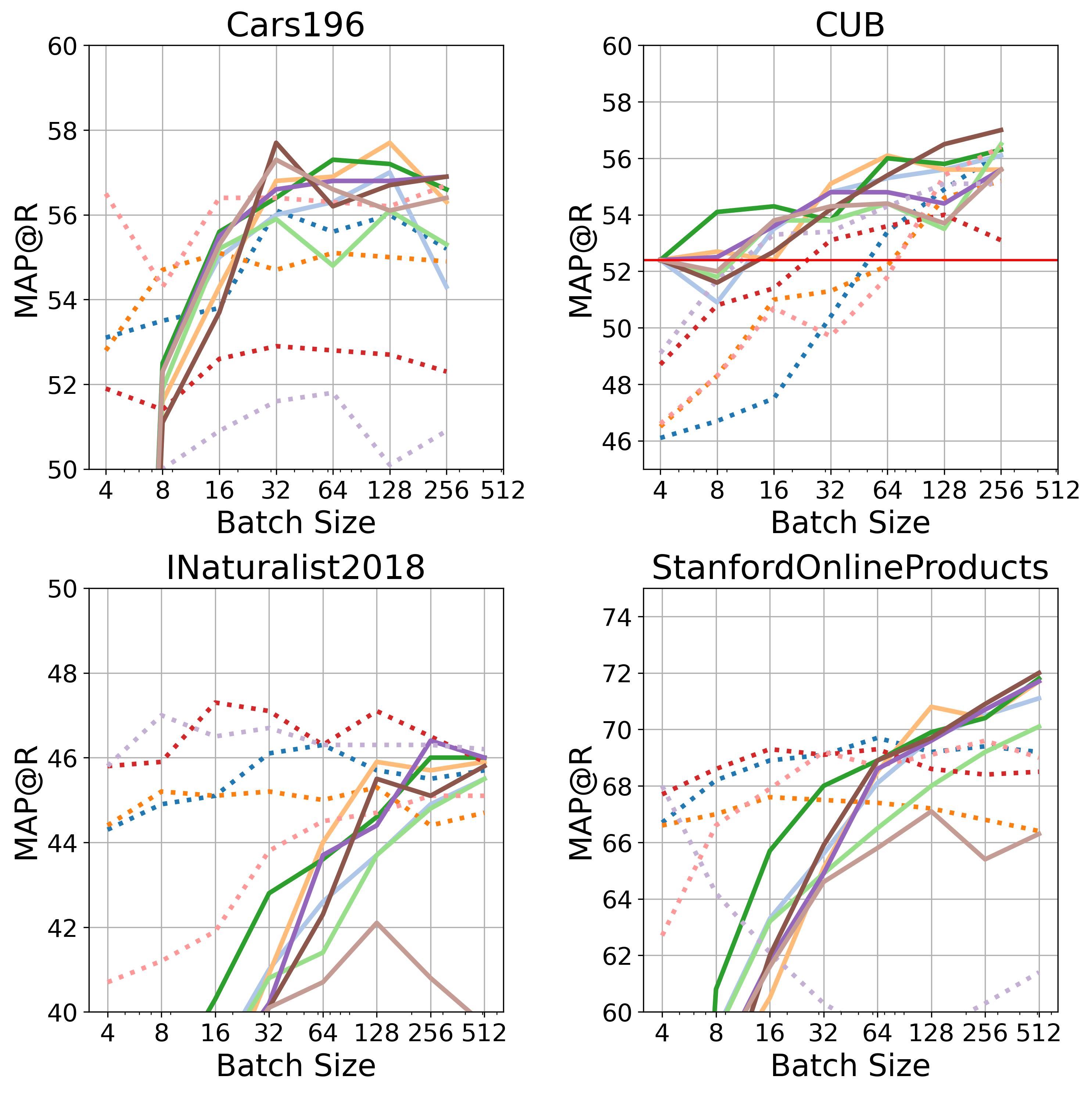}
    \end{center}
    \caption{\textbf{The accuracy of each loss function versus the batch size.} The red horizontal line represents the accuracy of the off-the-shelf DINO-v2; when not visible, the line is too low to be included in the plot.}
    \label{fig:b3_ablation_batch_size}
\end{figure}

A first question that naturally arises is: \textit{Should I use a different loss if I have lots of GPU memory, and therefore can afford large batch sizes?}
Hence, we ran experiments using different batch sizes, as shown in \cref{fig:b3_ablation_batch_size}. Here is a summary of the results:
\begin{itemize}
    \item On easy datasets (CUB, Cars) the choice of the loss does not matter much, and the results are noisy.
    \item Classification losses work well even with very small batch sizes.
    \item Contrastive losses greatly benefit from larger batch sizes, overtaking classification losses with batch sizes around 256-512.
    \item On the CUB dataset, the off-the-shelf model achieves similar results to the fine-tuned models, while on all other datasets, fine-tuning is crucial for good results.
\end{itemize}

\noindent Given these findings, we set the batch size to 256 for all the experiments in the upcoming sections.

\begin{figure}[h]
    \begin{center}
    \includegraphics[width=0.6\linewidth]{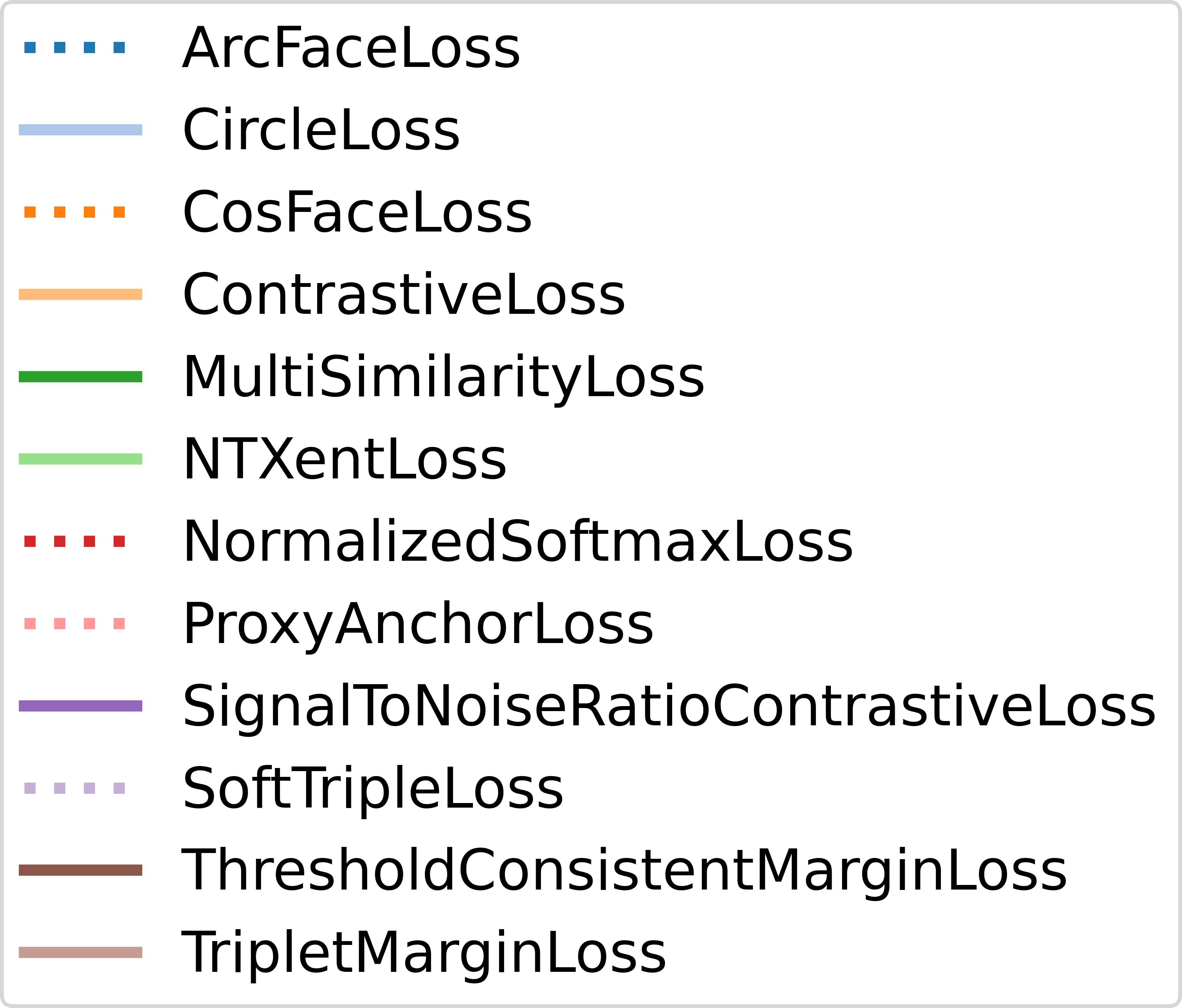}
    \end{center}
    \caption{\textbf{Legend of colors for each loss throughout each section of this paper.} A much smaller (almost unreadable) version of this legend is also shown in every plot.}
    \label{fig:legend}
\end{figure}

\subsection{Noisy / wrong labels}
\begin{figure}[h]
    \begin{center}
    \includegraphics[width=0.99\linewidth]{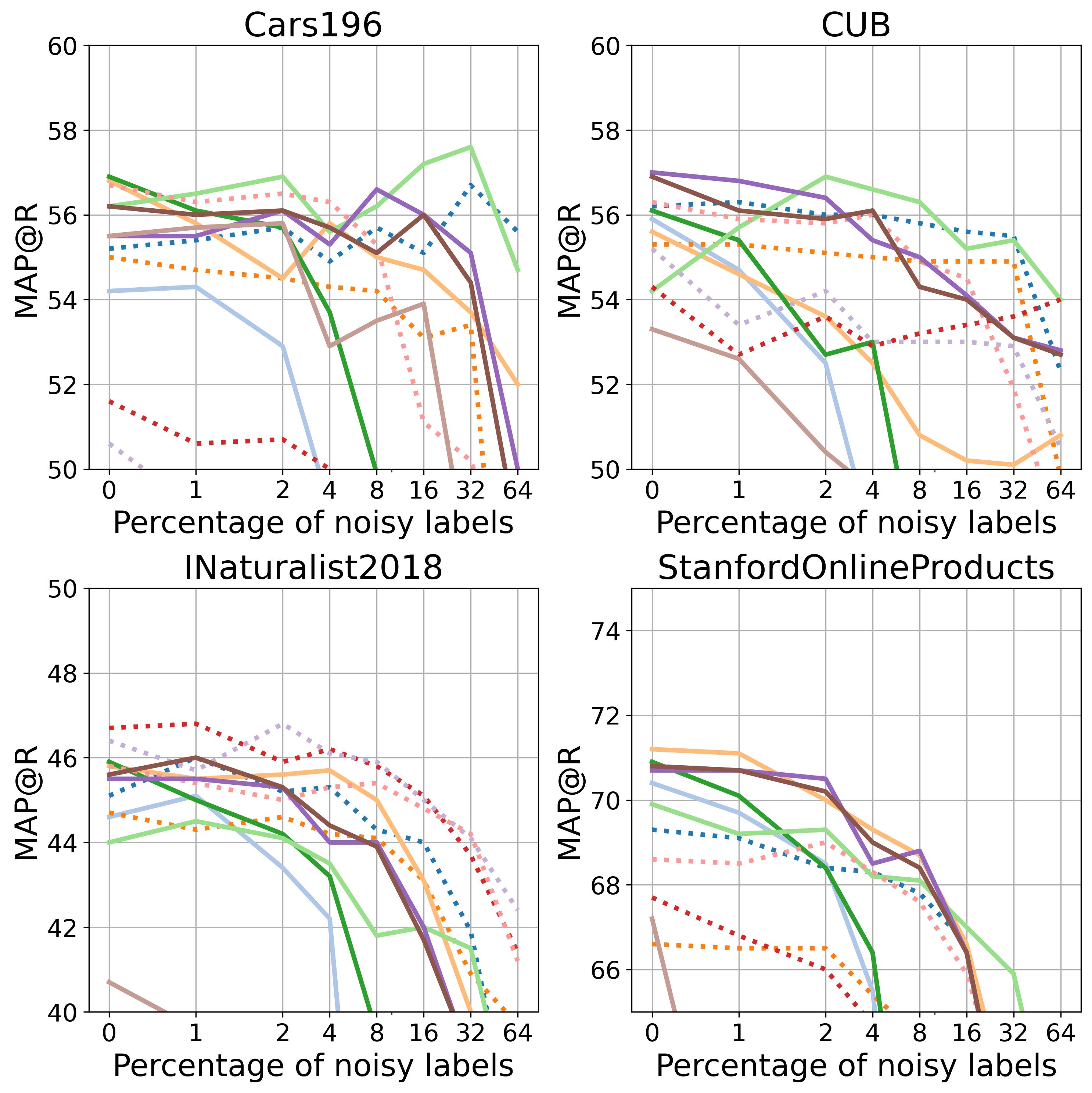}
    \end{center}
    \caption{\textbf{The accuracy of each loss function versus the percentage of data that is incorrectly labeled.} The x axis is the percentage of labels that are randomly changed during training.}
    \label{fig:b4_ablation_noisy_labels}
\end{figure}

When creating a training dataset, image retrieval developers might wonder: \textit{How much does annotation accuracy affect model performance?} To investigate this question, we conducted experiments by assigning different amounts of random labels to the training dataset. The results are shown in \cref{fig:b4_ablation_noisy_labels}.

Similarly to what is shown in \cref{fig:b3_ablation_batch_size}, we find that results on Cars196 and CUB are noisy, and surprisingly find that even high numbers (\eg 32\%) of wrong/random labels do not impact results much.
We also notice that, among the contrastive losses, the NTXentLoss is the most robust to high amounts of noisy labels.
Interestingly, we find that classification losses excel in iNaturalist2018, whereas contrastive losses perform better in SOP.

\subsection{Number of training classes}
In this section, we explore how accuracy is affected by the number of classes in the training set. (Note that the training, validation, and test sets do not share any classes.)

\begin{figure}[h]
    \begin{center}
    \includegraphics[width=0.99\linewidth]{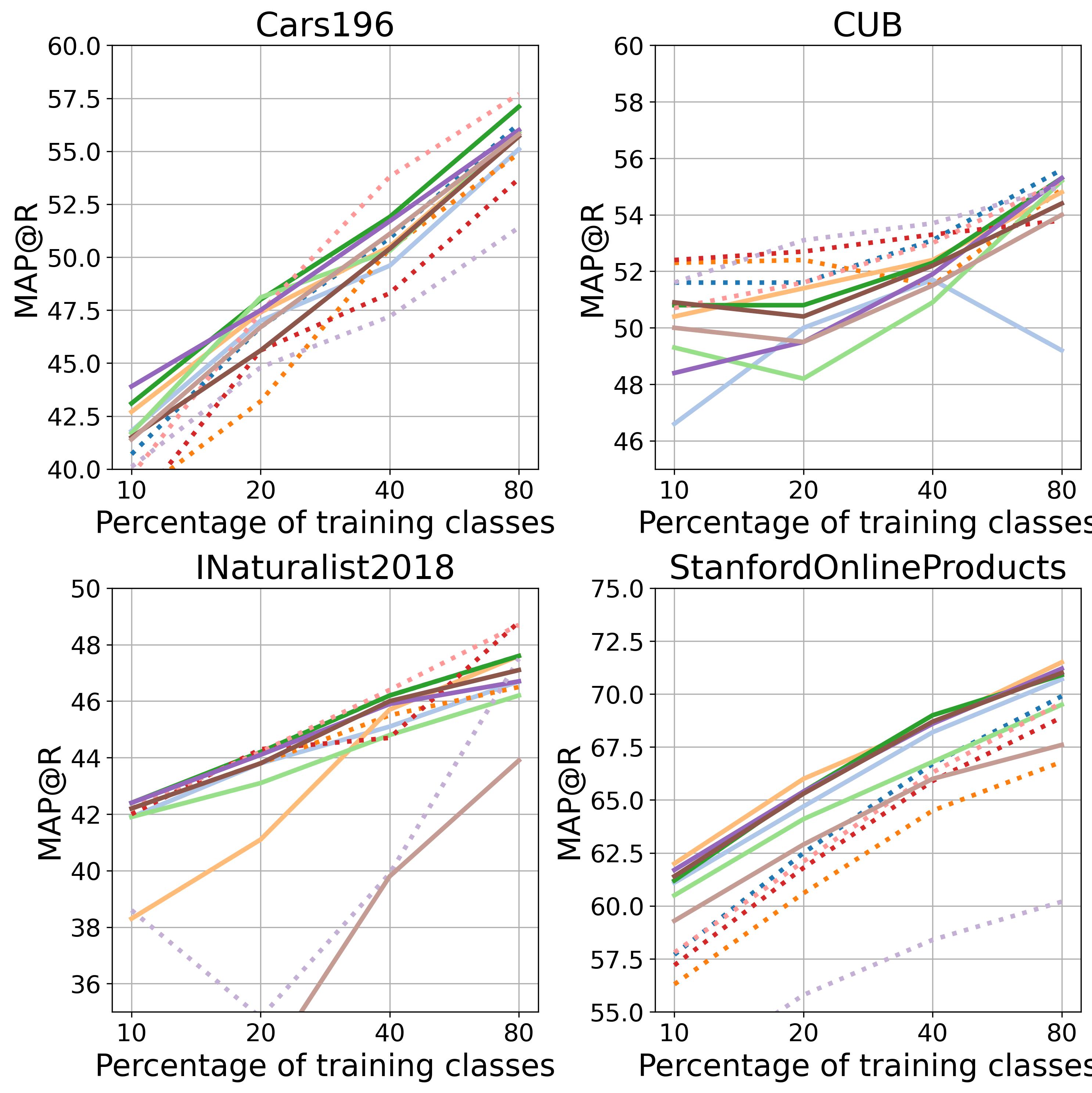}
    \end{center}
    \caption{\textbf{The accuracy of each loss function versus the percentage of training classes used.}}
    \label{fig:b5_ablation_train_classes}
\end{figure}

As shown in \cref{fig:b5_ablation_train_classes}, we reduce the number of classes in the training set by different amounts, and, somewhat unsurprisingly, we notice that this has a strong negative effect on accuracy. We also notice that the slopes of the accuracy lines are fairly uniform. In other words, when we reduce the number of training classes, all 12 loss functions show approximately the same pattern of accuracy decline, regardless of their initial accuracy levels.

\textit{So what does this mean in practice?}
While seemingly obvious, these results can have important implications in the real world.
Considering that one of the most important steps of an image retrieval application is labeling the data, these results, combined with our findings on noisy labels (\cref{fig:b4_ablation_noisy_labels}), suggest that when labeling images, one should not spend too much time making sure that each annotation is correct, but instead should focus on labeling \textit{more} data.

\subsection{Feature dimensionality}
One of the most important hyperparameters for an image retrieval model is the dimensionality of its output features (a.k.a ``embeddings" or ``descriptors'').
Generally, higher-dimensional features can encode more information, which should lead to better retrieval accuracy.

\begin{figure}[h]
    \begin{center}
    \includegraphics[width=0.99\linewidth]{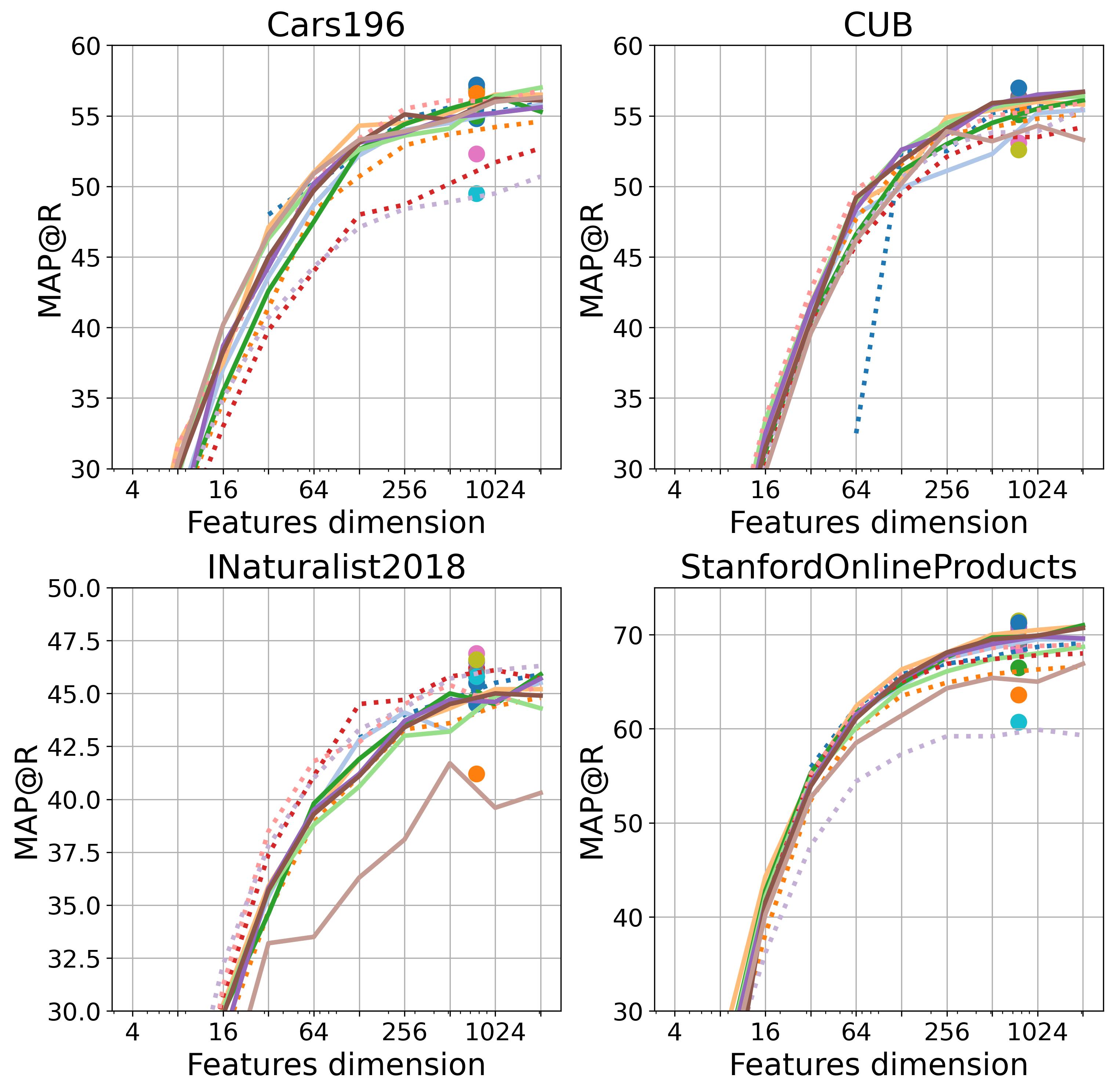}
    \end{center}
    \caption{\textbf{The accuracy of each loss function versus feature dimensionality.} Feature dimensionality is changed using a linear layer at the end of the model. The large dots represent the accuracy when no linear layer is used (\ie we directly use the features from the CLS token, which has dimensionality 768 for DINOv2-base).}
    \label{fig:b6_ablation_feat_dim}
\end{figure}

To understand how this relationship manifests across different loss functions, we run experiments with different feature dimensionalities, and report the accuracies in \cref{fig:b6_ablation_feat_dim}. Generally, we see that the best results are achieved when directly using the CLS token, which for DINOv2-base, is 768 dimensional.

It is important to keep in mind that the k-nearest-neighbors algorithm has linear complexity w.r.t. the dimensionality, both in terms of space and time.
While this is not an issue for small datasets, it can quickly become an issue for larger ones. As an example, given 1024-dimensional features, a dataset with 100 million images would need $100M \times 1024 \times 4 \sim 400GB$ of memory for its features (using common float32 implementations, \ie 4 bytes per value).

\subsection{Sampling}
\label{sec:sampler_m}
\begin{figure}[h]
    \begin{center}
    \includegraphics[width=0.99\linewidth]{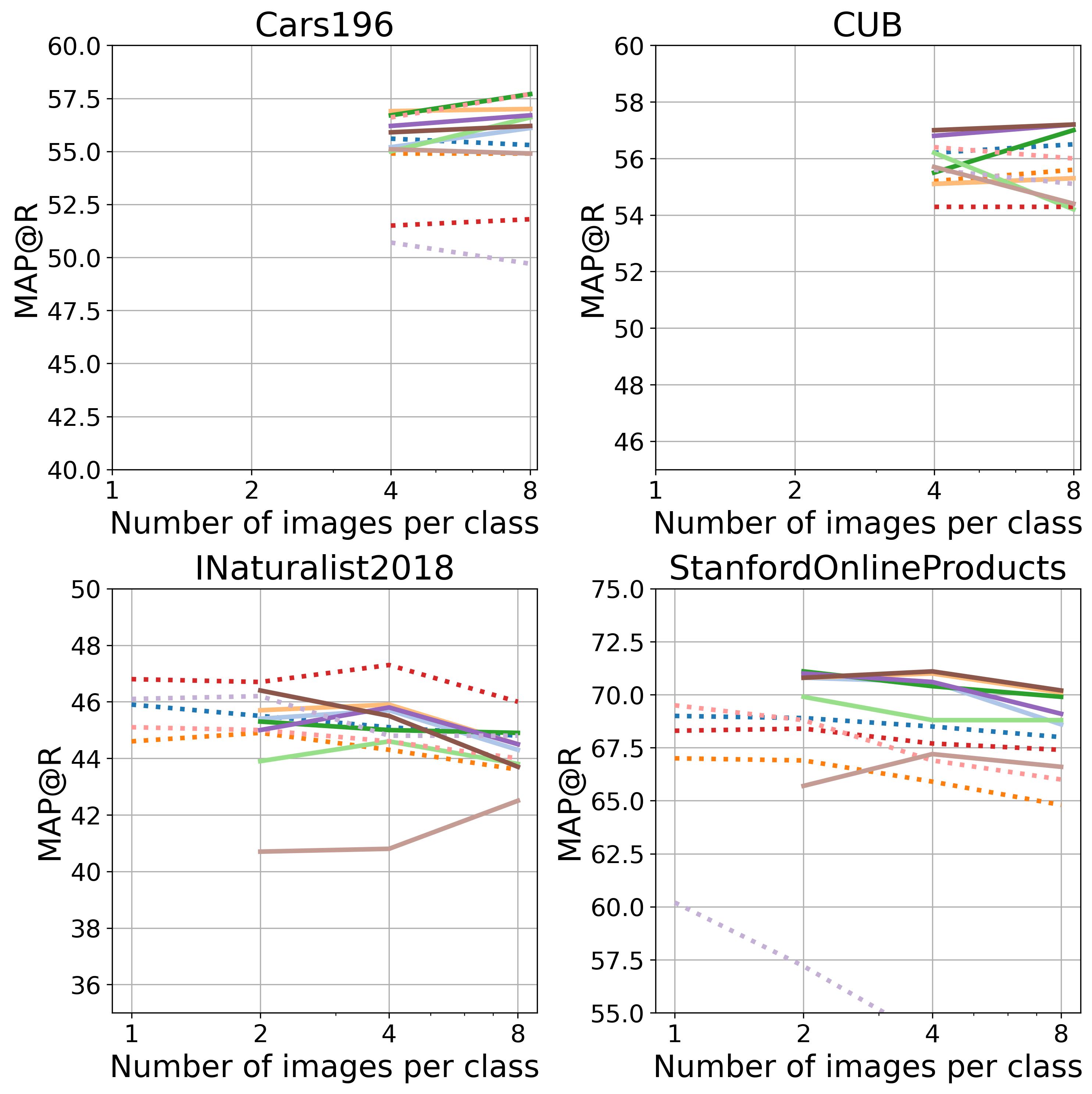}
    \end{center}
    \caption{\textbf{The accuracy of each loss function versus the number of samples per class ($M$) within a batch.} For example, if $M$ is 1 and the batch size is 256, each one of the 256 samples will come from a different class. Given that Cars196 and CUB have less than 128 classes, results with $M\leq2$ with batch size 256 cannot be computed (hence the missing lines). Similarly, contrastive losses cannot be computed with $M=1$ as they need in-batch positive samples, unlike classification losses.}
    \label{fig:b9_sampler_m}
\end{figure}

When creating a training batch, one important hyperparameter is the number of images sampled per class. Contrastive losses need at least two samples per class, so that positives are included in the batch, whereas classification losses have no such requirement. Let $M$ represent the value of this hyperparameter. Figure \ref{fig:b9_sampler_m} shows the relationship between $M$ and accuracy. Note that the batch size is fixed to 256, so when we select $M=4$, there will be 256 images from 64 different classes (4 per class).
Results show that classification losses are robust in the choice of $M$, although they generally benefit from using $M=1$, whereas contrastive losses fare well with $M=2$ and $M=4$, leading us to choose $M=4$ throughout the rest of the experiments.

\subsection{Learning Rate(s)}
\label{sec:lr}
\begin{figure}[h]
    \begin{center}
    \includegraphics[width=0.99\linewidth]{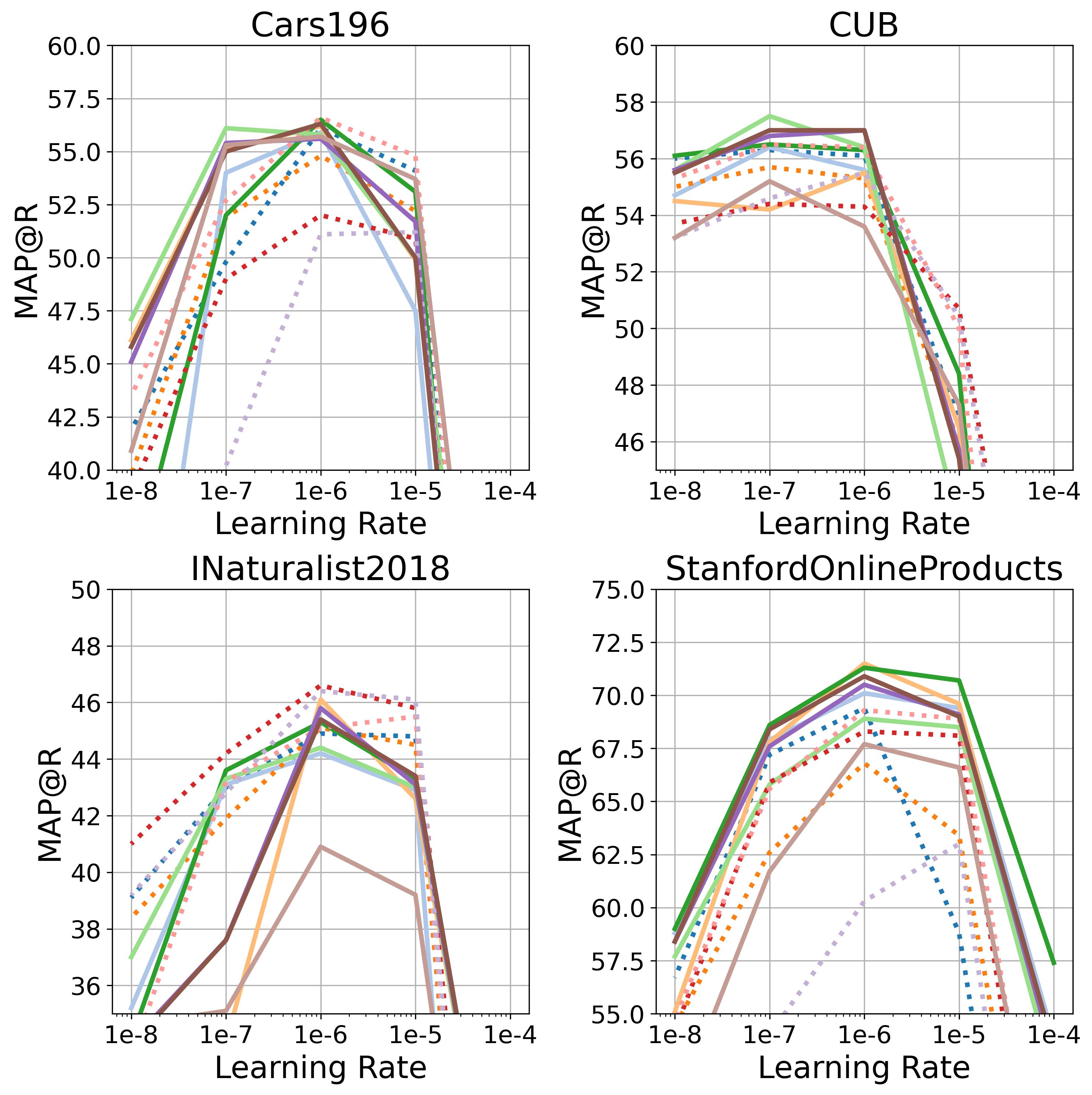}
    \end{center}
    \caption{\textbf{The accuracy of each loss function versus the learning rate of the model's optimizer.}}
    \label{fig:b7_ablation_lr}
\end{figure}

\begin{figure}[h]
    \begin{center}
    \includegraphics[width=0.99\linewidth]{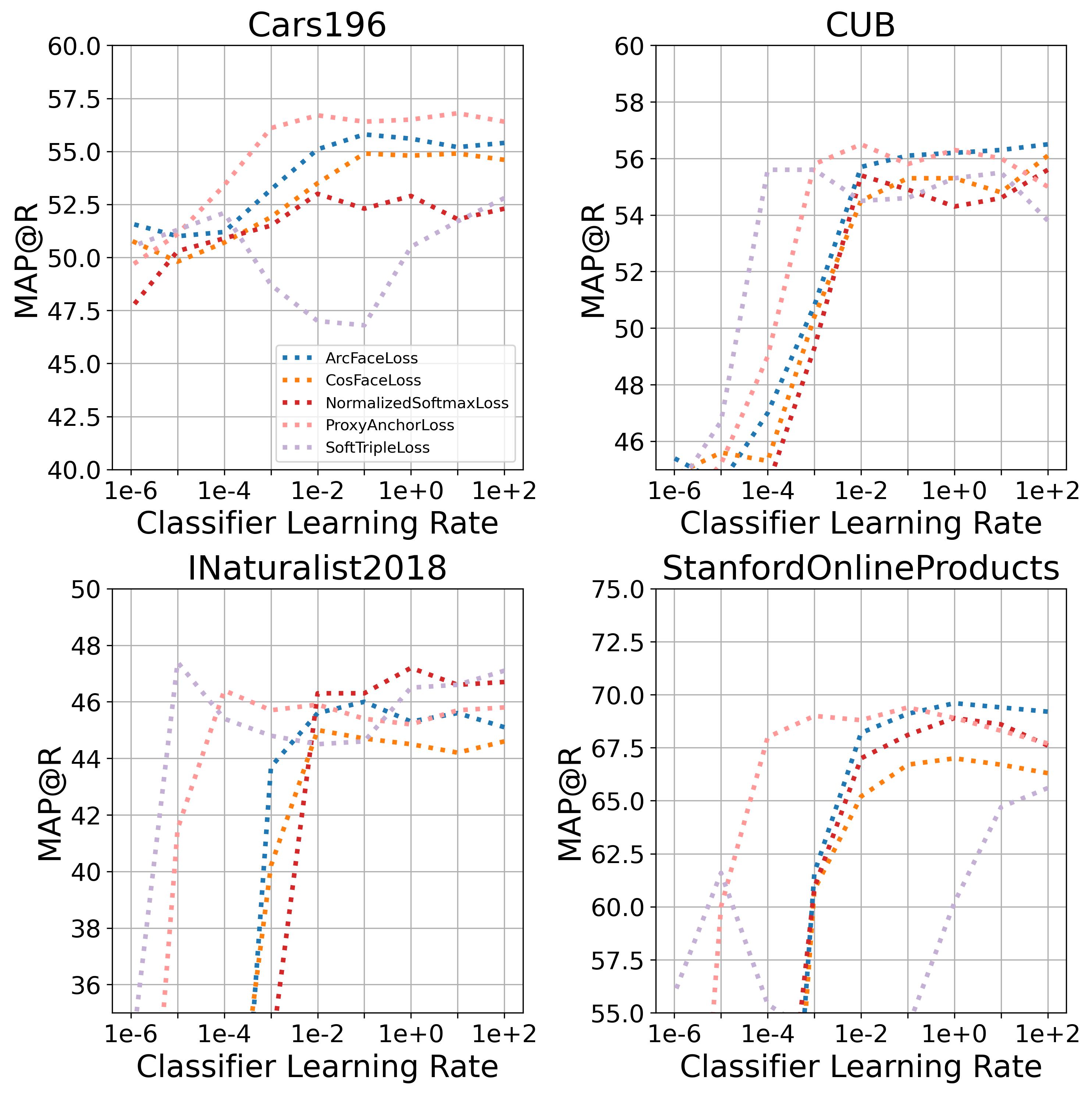}
    \end{center}
    \caption{\textbf{The accuracy of each loss function versus the learning rate of the classifier's optimizer.}}
    \label{fig:b8_ablation_classifier_lr}
\end{figure}

The learning rate is one of the most important training hyperparameters within a deep learning pipeline, as it strongly influences both the results and training speed.
With classifier losses, we find that it is a good practice to use two separate learning rates: one for the model, and one for the loss's parameters.
Figure \ref{fig:b7_ablation_lr} shows the effect of changing the learning rate for the model, and \cref{fig:b8_ablation_classifier_lr} shows the effect of changing the learning rate of the classifier/loss optimizer (hence only classification losses are shown).

The most notable insight is that the learning rate has a huge impact on the learning process, and choosing a sub-optimal learning rate has catastrophic consequences. On average, 1e-6 is a good choice for the model optimizer learning rate. 
Secondly, by comparing the two plots (\cref{fig:b7_ablation_lr} and \cref{fig:b8_ablation_classifier_lr}), we can see that it is very important to tune the learning rates for the model and classifier separately. In fact, using the same learning rate of 1e-6 for the classifier, often leads to a drop in MAP@R of over 10\%. Instead, the optimal classifier learning rate is around 1.0.
We also notice that the results from \cref{fig:b8_ablation_classifier_lr} are more stable than those in \cref{fig:b7_ablation_lr}, and that, for the classifier, choosing a high learning rate (\eg $>1$) is a safer choice than choosing a small one (\eg $<0.001$).
Finally, we emphasize that setting the wrong learning rate is something that could easily be overlooked, and can lead to disastrous results.

\section{Conclusion}

In this work, we explored the impact of various factors on the accuracy of image retrieval models, including the embedding model architecture, learning rates, batch size, loss function, data sampler, amount of label noise, and training dataset size. 
We found that the choice of loss function depends on the available resources: contrastive losses for large batch sizes and classification losses for smaller ones.  Additionally, our results suggest that focusing on labeling \textit{more} data, rather than ensuring high annotation quality, is more beneficial for training effective retrieval systems.
Finally, we found that it is crucial to tune the learning rate for the classifier separately from the model's learning rate.
These findings provide practical guidance for developing image retrieval systems, emphasizing the importance of balancing trade-offs between accuracy, computational resources, and data annotation strategies. 

{
    \small
    \bibliographystyle{ieeenat_fullname}
    \bibliography{main}
}

\end{document}